\documentclass[letterpaper]{article} 
\usepackage{aaai2026}  
\usepackage{times}  
\usepackage{helvet}  
\usepackage{courier}  
\usepackage[hyphens]{url}  
\usepackage{graphicx} 
\usepackage{natbib}  
\usepackage{caption} 
\usepackage{amsmath}
\usepackage{amsfonts}
\usepackage{arydshln}

\usepackage{natbib}  
\usepackage{caption} 
\usepackage{algorithm}
\usepackage{algorithmic}
\usepackage{booktabs}

\usepackage{newfloat}
\usepackage{listings}
\DeclareCaptionStyle{ruled}{labelfont=normalfont,labelsep=colon,strut=off} 
\floatstyle{ruled}
\newfloat{listing}{tb}{lst}{}
\floatname{listing}{Listing}
\title{PL-CA: A Parametric Legal Case Augmentation Framework}
\author {
    Ao Chang\textsuperscript{\rm 1,2},
    Yubo Chen\textsuperscript{\rm 1,2}\thanks{%
      Corresponding author, \protect\url{yubo.chen@nlpr.ia.ac.cn} },
    Jun Zhao\textsuperscript{\rm 1,2}
}
\affiliations {
    \textsuperscript{\rm 1}The Key Laboratory of Cognition and Decision Intelligence for Complex Systems,
Institute of Automation, Chinese Academy of Sciences, Beijing, China\\
    \textsuperscript{\rm 2}School of Artificial Intelligence, University of Chinese Academy of Sciences, Beijing, China\\
}

\usepackage{bibentry}

\begin{document}

\maketitle

\begin{abstract}

Conventional RAG is considered one of the most effective methods for addressing model knowledge insufficiency and hallucination, particularly in the judicial domain that requires high levels of knowledge rigor, logical consistency, and content integrity. However, the conventional RAG method only injects retrieved documents directly into the model's context, which severely constrains models due to their limited context windows and introduces additional computational overhead through excessively long contexts, thereby disrupting models' attention and degrading performance on downstream tasks. Moreover, many existing benchmarks lack expert annotation and focus solely on individual downstream tasks while real-world legal scenarios consist of multiple mixed legal tasks, indicating conventional benchmarks' inadequacy for reflecting models' true capabilities. 
To address these limitations, we propose PL-CA, which introduces a parametric RAG (P-RAG) framework to perform data augmentation on corpus knowledge and encode this legal knowledge into parametric vectors, and then integrates this parametric knowledge into the LLM's feed-forward networks (FFN) via LoRA, thereby alleviating models' context pressure. Additionally, we also construct a multi-task legal dataset comprising more than 2000 training and test instances, which are all expert-annotated and manually verified. We conduct our experiments on our dataset, and the experimental results demonstrate that our method reduces the overhead associated with excessively long contexts while maintaining competitive performance on downstream tasks compared to conventional RAG. Our code and dataset are provided in the appendix.

\end{abstract}

\section{Introduction}

In recent years, the rapid development of LLM in the field of natural language processing have led to their increasingly widespread applications in legal AI scenarios \cite{mullick2022evaluation, kim-etal-2024-self, fan2025legalruleinductiongeneralizable}. The legal tasks, including Legal Judgment Prediction (LJP), Statute Article Generation (SAR), Legal Document Generation (LDG), present greater challenges for the application of LLMs. Consequently, LLMs must accurately generate legal articles \cite{su-etal-2024-stard}, understand complex legal cases \cite{li2024lexevalcomprehensivechineselegal}, and make reasonable and accurate judgments \cite{he2024agentscourtbuildingjudicialdecisionmaking} across various legal tasks.

\begin{figure}[t]  
	\centering  
	\includegraphics[width=8cm, height=8cm]{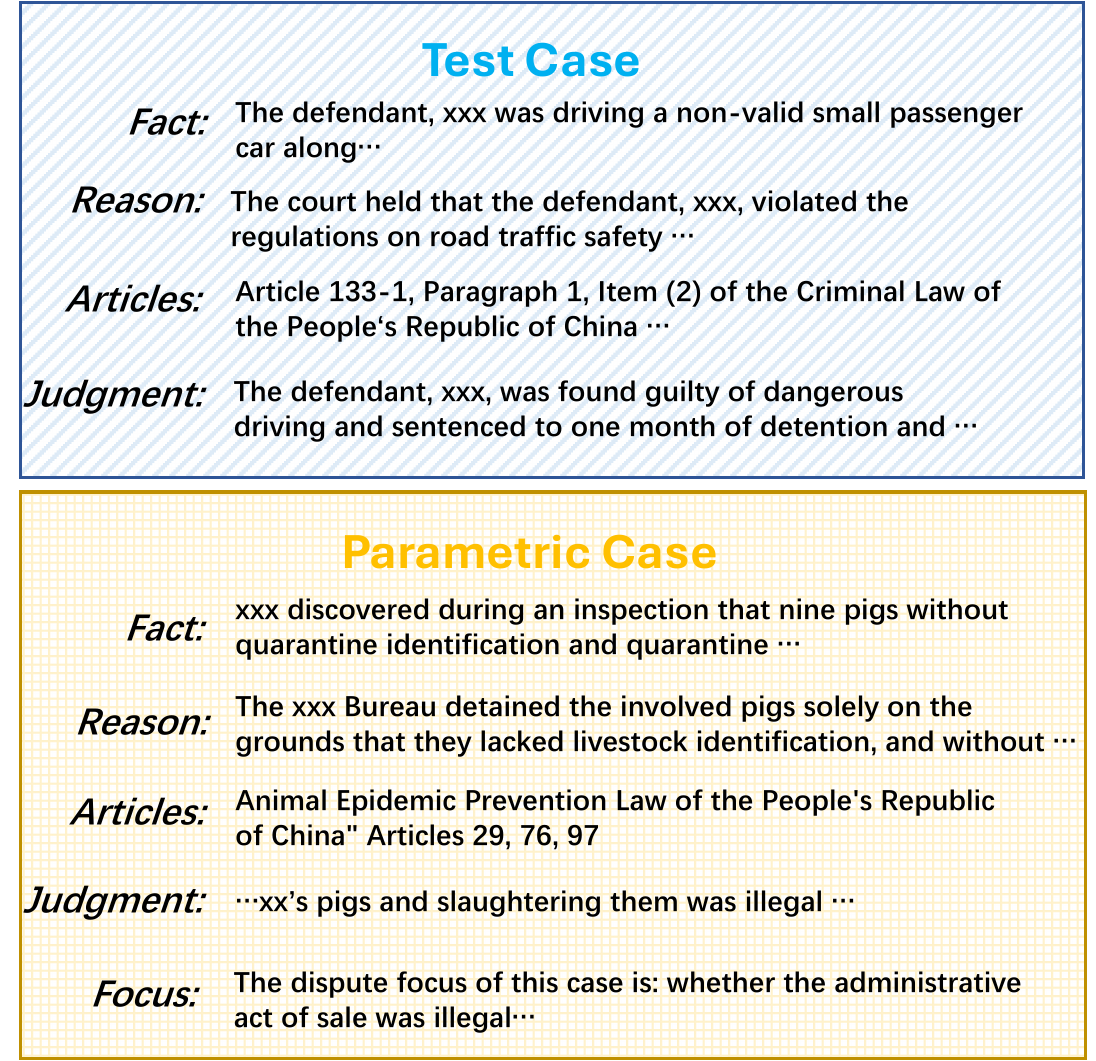}  
	\caption{The detailed structure of Test Cases and Parametric Cases, which is used for parametric injection.}
	\label{case_structure}  
\end{figure}

RAG has emerged as an effective strategy to mitigate the limitations of LLMs concerning knowledge insufficiency and hallucination issues by injecting relevant information from external knowledge bases into the model context \cite{li2024enhancingllmfactualaccuracy}. 
Traditional RAG provides additional knowledge support to the model by directly incorporating retrieved documents into its context. To further improve LLM's performance, some works \cite{li2024deltapretraindiscriminativeencoder, li2023sailerstructureawarepretrainedlanguage, xiao2021lawformerpretrainedlanguagemodel} focus on optimizing retrievers to recall more gold documents.  
However, despite its success in many domains, the limitations of RAG are becoming increasingly evident within the legal field. For example, legal texts often have exceptionally long contexts, and a single legal case may contain thousands of tokens on average \cite{li2023lecardv2largescalechineselegal}. Conventional RAG methods, which directly concatenate retrieved documents into the LLM's input context, present a significant bottleneck for models with limited context windows.  Too long contexts not only introduce substantial time and space overhead, thereby significantly increasing inference costs, but also negatively influence the model's attention to crucial information, consequently degrading performance on downstream tasks \cite{li2024longcontextllmsstrugglelong, qian2024longllmsnecessitylongcontexttasks}. More importantly, existing research indicates that LLMs generally perform better when utilizing its internal parametric knowledge \cite{yu2024neuronlevelknowledgeattributionlarge} than externally injected contextual knowledge.

\begin{table}[t]
\centering
\scalebox{0.8}{
    \begin{tabular}{lccc}
    \toprule
    \textbf{Dataset} & \textbf{SimuCourt} & \textbf{RareCases} & \textbf{Legal-CA} \\
    \midrule
    cases & 420 & 136 & 591 \\
    avg articles & 3.71 & 5.90 & 6.71 \\
    avg length per case fact & 346.1 & 491.3 & 605.8 \\
    avg token length & 3326 & 5807 & 415 \\
    fine-tune? & No & Yes & Yes \\
    \bottomrule
    \end{tabular}
}
\caption{Comparison with other datasets}
\label{datasets_difference}
\end{table}

Besides, current benchmarks \cite{li2025casegenbenchmarkmultistagelegal, fan2025legalruleinductiongeneralizable, su-etal-2024-stard, sun2024lawluomultiagentcollaborativeframework,chen-etal-2025-slard} in the legal domain often lack high-quality expert annotations and tend to focus solely on individual downstream legal tasks. However, real-world legal scenarios typically involve multiple interrelated tasks, making conventional benchmarks inadequate to fully reflect the true capabilities of LLMs in complex legal environments, thus failing to meet the demands of practical legal applications.

Inspired by P-RAG \cite{su2025parametricretrievalaugmentedgeneration}, Parametric Retriever Augmentation Generation,  we propose PL-CA: a \textbf{P}arametric \textbf{L}egal \textbf{C}ase \textbf{A}ugmentation framework to overcome the limitations of traditional RAG in the legal domain and address the deficiencies of current benchmarks. PL-CA introduces a parametric RAG approach that augments legal corpora with enhanced knowledge. This knowledge is then efficiently integrated into the LLM using Low-Rank Adaptation (LoRA) techniques. This approach effectively reduces the burden on the model's context window, enabling it to internally encode more legal knowledge and thereby improving its ability to utilize such knowledge. 

Furthermore, to enable a more comprehensive and accurate evaluation of LLM performance in the legal domain, we construct Legal-CA, a multi-task legal dataset. This dataset comprises 590 expert-annotated and manually verified test samples and 1,990 training samples, covering three legal areas, criminal law, administrative law, and civil law, and encompasses various downstream legal tasks, which is designed to better reflect model performance in realistic, complex legal scenarios.

Experimental results show that our PL-CA method reduces the computational cost caused by excessively long contexts without compromising performance on downstream tasks compared to conventional RAG approaches. Particularly in tasks requiring deep understanding and reasoning of legal knowledge—such as legal article retrieval and judgment prediction—PL-CA shows significant advantages. These findings strongly validate the effectiveness and feasibility of P-RAG in the legal domain, offering new insights and methodologies for the application of LLMs in legal practice.

Our contributions are summarized as follows:
\begin{itemize}
\item[$\bullet$] We have proposed the PL-CA framework, which is the first to apply P-RAG to legal downstream tasks, alleviate the issues in some degree caused by excessively long legal texts in past tasks.
\item[$\bullet$] We construct Legal-CA, a legal dataset comprising 590 test instances and 1990 train instances, along with Legal-KD, a legal knowledge case corpus, to assess LLMs' legal capacity comprehensively.
\item[$\bullet$] We conduct extensive experiments to validate the effectiveness of our methods. Our approach enables the LLM to achieve significant improvements on downstream tasks, even surpassing GPT-4o in certain domains.
\end{itemize}

\begin{table}[t]
\centering
\scalebox{0.8}{
    \begin{tabular}{lccc}
    \toprule
    \textbf{Dataset} & \textbf{civil} & \textbf{crime} & \textbf{admin} \\
    \midrule
    number of cases & 17.55M & 2.45M & 0.87M \\
    \bottomrule
    \end{tabular}
}
\caption{Statistic of Legal-KD}
\label{legal-kd}
\end{table}

\section{Related Work}
Many works have emerged to assess models' capacity to handle legal problems. To assess models' LJP ability, several datasets like CAIL\cite{xiao2018cail2018largescalelegaldataset} have been proposed. 
Subsequently, with the advent of LLMs, several legal benchmark datasets have been subsequently proposed to assess LLMs' capabilities in various downstream legal tasks, including LegalBench\cite{pipitone2024legalbenchragbenchmarkretrievalaugmentedgeneration} from the United States, LawBench\cite{fei2023lawbenchbenchmarkinglegalknowledge} from China, and LexFile\cite{chalkidis2023lexfileslegallamafacilitatingenglish} from Europe. 

Consequently, researchers begin to explore LLMs' legal capabilities in alternative scenarios. In the LJP task, \cite{wu2023precedentenhancedlegaljudgmentprediction} enhanced LLM performance on downstream tasks by fine-tuning models with legal precedents.
In the Legal Case Retrieval (LCR) domain, LeCard\cite{li2023lecardv2largescalechineselegal} is proposed as a LCR dataset to evaluate retrieval models' capacity. GEAR\cite{Qin_2024} integrates LJP and LCR tasks by constructing document trees to improve recall capabilities for legal cases. Furthermore, \cite{gao2024enhancinglegalcaseretrieval, zhang2025citalawenhancingllmcitations} optimize downstream task performance through enhanced retrieval mechanisms, while \cite{sun2024lawluomultiagentcollaborativeframework} and \cite{barron2025bridginglegalknowledgeai} adopt knowledge graph approaches to enhance legal corpus retrieval capabilities.
For the Statute Article Retrieval (SAR) task, STARD\cite{su-etal-2024-stard} constructed a comprehensive statute knowledge base, including datasets and corpora. In contrast, SLARD\cite{su-etal-2024-stard} focuses on municipal regulations rather than statutory articles and proposes a new benchmark suite. Across these retrieval-oriented downstream tasks, model performance remains generally suboptimal, revealing the gap between models' legal knowledge comprehension and application capabilities.

In the realm of agent-based legal applications, LLMs have also demonstrated significant contributions. AgentCourt\cite{chen2025agentcourtsimulatingcourtadversarial} enhances legal knowledge by utilizing debate information generated by lawyer agents as a corpus, enabling models to retrieve it when addressing downstream tasks. SimuCourt\cite{he2024agentscourtbuildingjudicialdecisionmaking} proposes new datasets and extensive case data based on the debate framework. ASP2LJ\cite{chang2025asp2ljadversarialselfplay} improves agent debating capabilities by generating synthetic case data, mitigating limitations arising from insufficient existing legal cases and long-tail distribution issues. 
MASER\cite{yue2025multiagentsimulatordriveslanguage} presents a more realistic scenario by not only simulating multi-turn question-answering interactions in legal consultation but also imparting clients with diverse personalities based on the Big-5, thereby generating more realistic and complex data.

Furthermore, beyond conventional tasks such as LJP and LCR, researchers are expanding into novel legal scenarios to comprehensively evaluate LLMs' legal capabilities. \cite{li2025casegenbenchmarkmultistagelegal} proposes the CaseGen benchmark to test models' legal document generation abilities, while JuDGE\cite{su2025judgebenchmarkingjudgmentdocument} integrates LJP with document generation to analyze model performance across multiple dimensions. To better approximate the complexity of real-world cases, \cite{li2024legalagentbenchevaluatingllmagents} introduces LegalAgentBench, which analyzes LLM capabilities across 17 realistic legal scenarios and 37 external knowledge base interaction tools. In legal consultation tasks, LexRAG proposes a multi-turn legal consultation framework and organizes it into an RAG toolkit, empowering subsequent legal RAG research.

\begin{figure}[t]  
	\centering  
	\includegraphics[width=8cm, height=6cm]{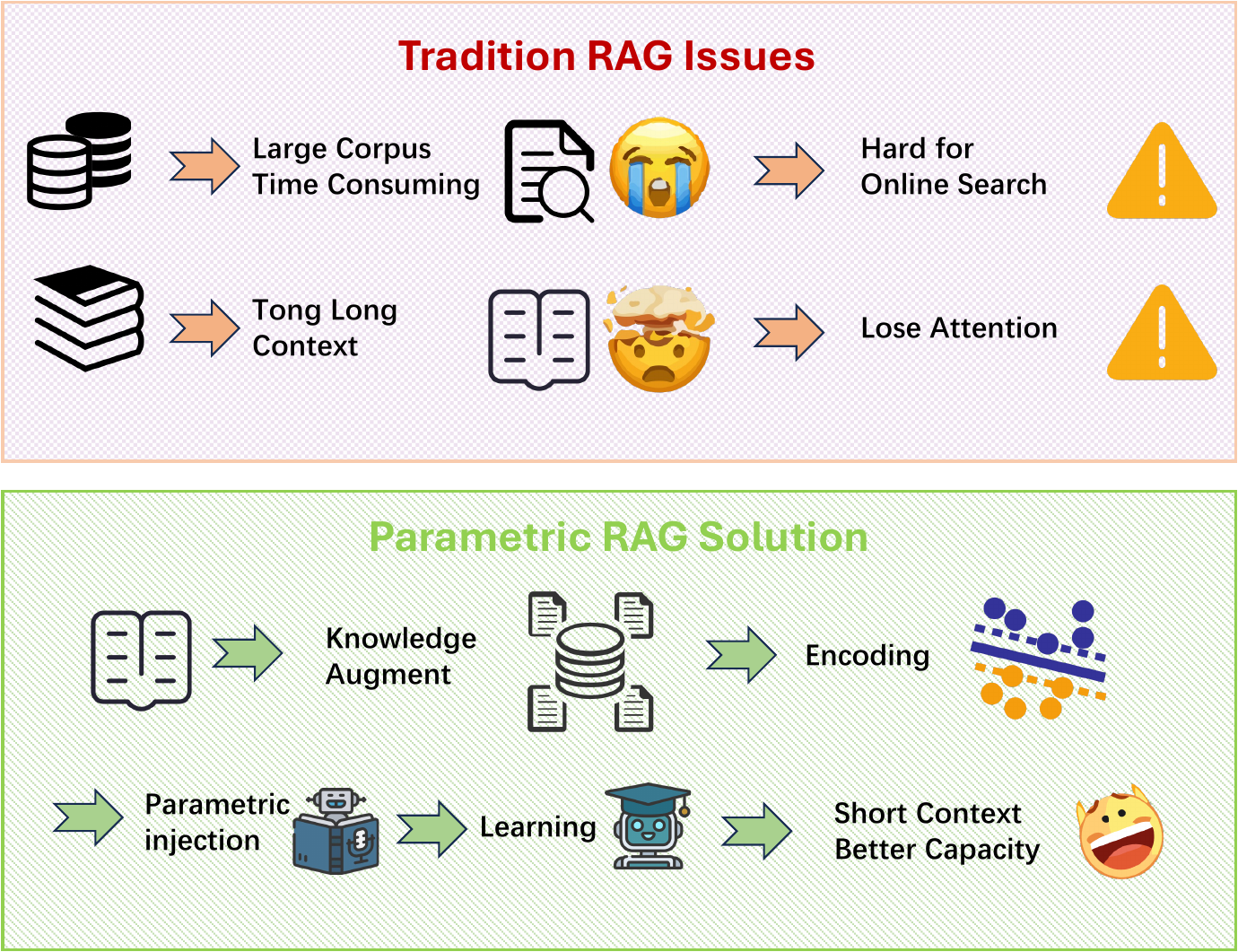}  
	\caption{Comparison between traditional RAG and P-RAG.}
	\label{Comparison}  
\end{figure}

\section{Methodology}
\subsection{Preliminary}

As illustrated in Figure~\ref{case_structure}, a complete legal judgment document typically comprises four main components: case facts, legal reasoning, judgment results, and relevant statutes. Correspondingly, typical legal judgment scenarios involve three primary downstream tasks: LJP, SAG, and LDG. These tasks are sequentially organized to produce legal responses based on the provided factual information.


To evaluate the performance of LLMs on these tasks, we define them as follows: given a case fact \textit{f}, an LLM $\Theta$ is expected to generate a response $\textit{r} = \Theta(\textit{f})$. 
In general, relying solely on the internal legal knowledge and reasoning capabilities of LLMs often fails to produce satisfactory results. Therefore, external knowledge is typically incorporated via RAG to enhance performance.

\begin{table}[t]    
    \centering
    \begin{small}
    \begin{tabular}{lccc}
        \toprule
        Model(\%) & Recall@5 & Recall@20 & Recall@100\\
        \midrule
        BM25 & 3.60 & 6.57 & 11.62 \\
        LawFormer & 0.08 & 0.26 & 1.20 \\
        bge-m3 & 1.39 & 3.58 & 9.04 \\
        ChatLaw-Text2Vec & 2.24 & 6.44 & 10.16 \\
        SAILER\_zh & 0.18 & 0.39 & 1.24 \\
        \bottomrule
    \end{tabular}
    \end{small}
    \caption{Different retrievers' article recall in Legal-CA,}
    \label{article_recall}
\end{table}

\subsubsection{Vanilla RAG.} 
For legal retrieval tasks, this approach uses the basic case fact $f$ as the input query. A sparse or dense retriever performs similarity matching between $f$ and a corpus $\mathcal{D} = \{d_1, d_2, \dots, d_n\}$, selecting the top-$k$ documents as the retrieved set $D = \{d'_1, d'_2, \dots, d'_k\}$. The query $f$ and the retrieved documents $D$ are then concatenated and provided as contextual input to the LLM. The model subsequently generates the final output as $\textit{r} = \Theta(f, D)$.

While this approach is conceptually simple and easy to implement, it often leads to performance degradation when handling lengthy legal documents, due to the input length limitations and context dilution in LLMs.

\subsubsection{P-RAG.} 
\citet{su2025parametricretrievalaugmentedgeneration} proposes a novel RAG paradigm, P-RAG, which enhances LLMs' ability to comprehend and utilize knowledge by injecting external knowledge directly into model parameters. As illustrated in Figure~\ref{Comparison}, we employ P-RAG to improve LLMs' performance, which is better than traditional RAG.

\begin{equation}
\label{eq:delta-theta-P-RAG}
\begin{split}
\Delta \Theta &= \text{Encode}\left(\{d_i\}_{i=1}^k\right), \\
\Theta' &= \Theta + \Delta \Theta.
\end{split}
\end{equation}

As shown in Equation~\ref{eq:delta-theta-P-RAG}, the retrieved documents are encoded and used to update the parameters of the LLM, thereby internalizing external knowledge. Recent studies~\cite{yu2024neuronlevelknowledgeattributionlarge} have demonstrated that LLMs are generally more effective at leveraging internalized knowledge than contextually provided external information. 

In the legal domain, where accurate and efficient legal reasoning is critical, this characteristic is particularly valuable. To improve LLM performance in legal tasks, we therefore propose to augment and parameterize legal knowledge. Following the design in~\citet{su2025parametricretrievalaugmentedgeneration}, we divide the procedure into two stages: offline and online.

However, given the large scale of the legal corpus, fully parameterizing the entire dataset would introduce significant computational overhead. Thus, in the offline stage, we select 1,990 representative legal samples from the authoritative legal website PKULaw to construct the offline corpus $\mathcal{D}_{\text{offline}}$. For the online stage, we introduce a new corpus, Legal-KD, as the online knowledge base $\mathcal{D}_{\text{online}}$.

\begin{figure*}[t]  
	\centering  
	\includegraphics[width=14cm, height=6cm]{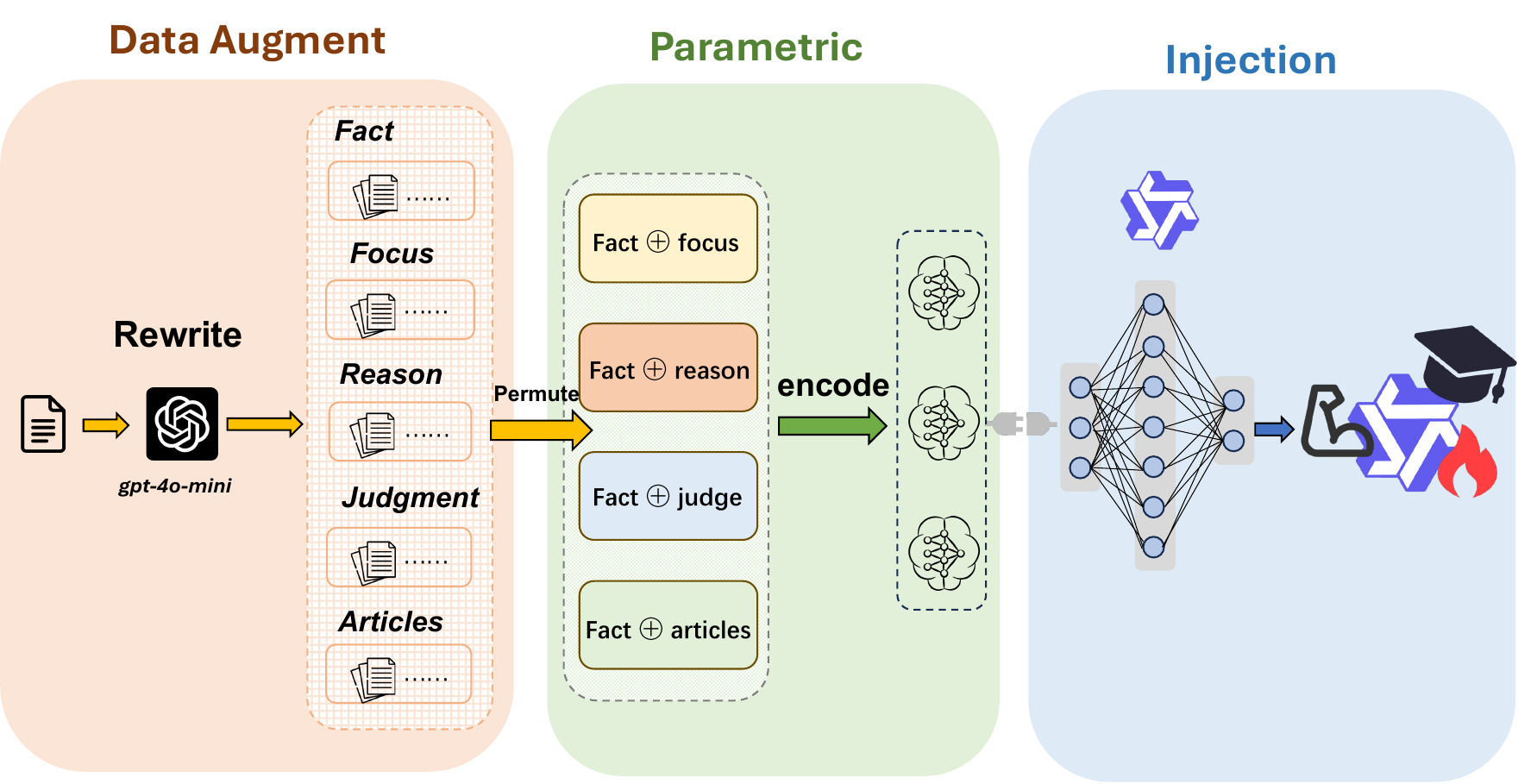}  
	\caption{Overview of the Case Augmentation and LLM Injection Pipeline. Our method involves five document sections: fact, focus, reason, judgment, and article. Each section is rewritten by GPT-4o-mini for case augmentation. The fact section is then concatenated with the other augmented sections, and the combined item is encoded into embeddings. Finally, LoRA is employed to inject these embeddings into the LLM.}
	\label{pipeline}  
\end{figure*}

\subsection{Offline P-RAG}

Given an LLM $\Theta$, an offline corpus $\mathcal{D}_{\text{offline}}$, and a set of queries $\mathcal{Q} = \{q_1, q_2, \dots, q_n\}$, where each query $q_i$ is paired with a corresponding answer $a_i$, the dataset can be represented as $(\mathcal{Q}, \mathcal{A}) = \{(q_i, a_i)\}$.

The structure of each legal case in the offline corpus is illustrated in Figure~\ref{case_structure}. To improve the model’s generalization capability with respect to legal knowledge, we perform data augmentation on each component of the case structure, including: case facts, legal reasoning, dispute focus, applicable statutes, and judgment outcome. In this setup, the case fact is treated as the query, while the remaining components serve as corresponding answers.

For the data augmentation process, we employ GPT-4o-mini to generate diverse variants of each data point. Each sample is rewritten three times, ensuring that the core legal information remains unchanged—this includes charges, sentencing outcomes, fines, and other legally significant details (e.g., names, locations, and terminology). As a result, the size of each component increases by a factor of four, and the number of QA pairs per case is expanded by a factor of sixteen (4 components × 4 total variations), thus significantly enhancing the model’s ability to internalize and reason about legal knowledge. The augmentation process for each component is detailed as follows:

\begin{itemize}
    \item \textbf{Case Fact}: Preserve charges, sentencing, and key factual details (e.g., monetary amounts, locations), while rewriting the text using alternative legal terminology.
    \item \textbf{Reason}: Rephrase sentence structures without altering the underlying legal logic (e.g., rewriting “constitutes theft” as “meets the constitutive requirements of theft”).
    \item \textbf{Focus}: Substitute key dispute phrases with synonymous legal expressions (e.g., “whether it constitutes voluntary surrender” as “whether it satisfies the condition of voluntary surrender”).
    \item \textbf{Article}: Retain statute numbers while paraphrasing the legal provisions using varied expressions.
    \item \textbf{Judgment}: Keep the judgment outcome consistent, but vary the phrasing (e.g., “sentenced to three years’ imprisonment” as “sentenced to imprisonment for a term of three years”).
\end{itemize}

\begin{table*}[t]
\centering
\normalsize
\begin{tabular}{lcccccccccc}
\toprule
\textbf{Model} & LA-P & LA-R & LA-F1 & Charge & Imprison & Probation & Fine & CA-P & CA-R & CA-F1  \\
\midrule
LexiLaw & 15.82 & 9.92 & 12.19 &  80.91 & 37.95 & 38.58 & 37.25 & 3.26 & 3.66 & 3.45 \\
ChatLaw & 8.91 & 5.31 & 6.66 &  73.91 & 27.92 & 38.95 & 30.25 & 7.08 & 9.04 & 7.94 \\
gpt-4o-mini & 4.45 & 4.32 & 4.38 &  66.85 & 38.82 & 36.59 & 39.21 & 9.85 & 13.18 & 11.27 \\
gpt-3.5-turbo-0125 & 4.19 & 3.39 & 3.75 &  78.42 & 34.75 & 37.53 & 37.17 & 14.22 & 23.29 & 17.66 \\
gpt-4o-2024-11-20 & 17.28 & 13.94 & 15.43 &  \underline{88.62} & \textbf{43.42} & \textbf{40.79} & \textbf{43.63} & \underline{25.75} & \textbf{31.54} & \textbf{28.35} \\
Qwen1.5-7B-Chat & 15.3 & 10.78 & 12.65 &  78.92 & 37.62 & 38.45 & 37.61 & 4.45 & 7.36 & 5.54 \\
Qwen1.5-7B-Chat+RAG & 20.35 & 13.96 & 16.56 &  87.78 & 39.67 & 38.85 & 40.12 & 10.48 & 15.46 & 12.49 \\
\hdashline
PL-CA & \underline{33.14} & \textbf{18.83} & \textbf{24.01} &  \textbf{92.39} & \underline{43.40} & \underline{40.52} & \underline{41.53} & \textbf{27.53} & \underline{27.06} & \underline{27.29} \\
Combine Both & \textbf{48.25} & \underline{14.41} & \underline{22.19} &  87.22 & 40.57 & 38.73 & 40.38 & 16.63 & 18.49 & 17.51 \\
\midrule
\end{tabular}
\caption{Fine-grained performance of Legal-CA. The prefix "LA" means legal article, "CA" means Civil \& Admin.}
\label{fine-grained}
\end{table*}

\begin{table*}[t]
\centering
\normalsize
\begin{tabular}{lcccccc}
\toprule
\textbf{Model} & J-P & J-R & J-F1 & R-P & R-R & R-F1 \\
\midrule
LexiLaw & 62.96 & 65.68 & 64.10 & 44.29 & 72.37 & 54.95 \\
ChatLaw & 64.63 & 73.03 & 68.34 & 39.53 & 74.51 & 50.74 \\
gpt-4o-mini & 61.69 & 78.06 & 68.67 & 57.43 & 79.46 & 66.24 \\
gpt-3.5-turbo-0125 & \underline{65.02} & 75.32 & 69.61 & 47.22 & 78.51 & 57.44 \\
gpt-4o-2024-11-20 & 55.06 & \underline{81.49} & 65.11 & \textbf{76.99} & \textbf{76.01} & \textbf{76.75} \\
Qwen1.5-7B-Chat & 63.94 & 79.42 & \underline{70.67} & 53.72 & 74.15 & 63.26 \\
Qwen1.5-7B-Chat+RAG  & 48.86 & 80.10 & 57.34 & 53.43 & 77.86 & 62.91 \\
\hdashline
PL-CA(P-RAG) & \textbf{76.01} & 72.52 & \textbf{73.94} & \underline{72.76} & \underline{75.48} & \underline{73.52} \\
Combine Both & 37.91 & \textbf{82.48} & 50.14 & 72.27 & 75.03 & 72.83 \\
\midrule
\end{tabular}
\caption{Coarse-grained performance of Legal-CA. The prefix "J" means judgment, and "R" means reason.}
\label{coarse-grained}
\end{table*}

After obtaining the augmented dataset $\mathcal{D}' = \{(q', a')\}$, we proceed with parameterized knowledge injection into the LLM. Specifically, we adopt Low-Rank Adaptation (LoRA) and tailor it to the legal domain to construct the PL-CA pipeline. The procedure consists of the following steps:

\begin{itemize}
    \item \textbf{LoRA Parameter Initialization}: For each legal case $d_i \in \mathcal{D}'$, we initialize a pair of low-rank matrices $A_i \in \mathbb{R}^{h \times r}$ and $B_i \in \mathbb{R}^{k \times r}$, where $h$ is the hidden dimension, $k$ is the feed-forward network (FFN) intermediate dimension, and $r \ll \min(h, k)$ denotes the LoRA rank.

    \item \textbf{Training Objective}: For each QA pair $(q', a') \in \mathcal{D}'$, we concatenate the question and answer into an input sequence $x = [q' \oplus a']$, where $\oplus$ denotes token-level concatenation. We use the standard next-token prediction objective to optimize the trainable LoRA parameters. Let $\Theta$ denote the frozen parameters of the pre-trained LLM, and $\Delta \Theta_i = \{A_i, B_i\}$ represent the trainable LoRA adapters. The training objective is defined as:

    \begin{equation}
    \label{equation2}
    \min_{\Delta \Theta_i} \sum_{(q', a') \in \mathcal{D}'_i} \sum_{t=1}^{T} -\log P_{\Theta + \Delta \Theta_i} (x_t \mid x_{<t})
    \end{equation}

    \item \textbf{Parameter Merging}: After training, each legal case $d_i$ is associated with a specific set of LoRA parameters $\Delta \Theta_i = \{A_i, B_i\}$. These parameters are stored as the case-specific, parameterized representation of the legal knowledge embedded in $d_i$.
\end{itemize}

\subsection{Online P-RAG}

After completing the offline preprocessing of the corpus, the model acquires a foundational level of legal knowledge. Similar to the offline PRAG method, the online stage also performs data augmentation, with the key distinction being the incorporation of a retrieval component.

\textbf{Retrieval}: The retriever performs online retrieval over the online corpus, selecting the top-1 most relevant case and extracting the corresponding legal statutes from the top-5 retrieved cases. These are used as the basis for subsequent online parameterized injection.

\textbf{Parametric Injection}: Following the parameterization strategy from the offline stage, the retrieved case and legal articles are structurally reformatted. An example of the online case structure is illustrated in Figure~\ref{case_structure}. It is worth noting that, due to structural differences between the offline and online corpora, the offline corpus is decomposed into four components: fact, reasoning, judgment, and applicable statutes. After applying data augmentation, the parameterized update is carried out according to Equation~\ref{equation2}, and the resulting parameters are injected into the LLM’s internal representation.

\begin{equation}
\label{eq:delta_theta}
\Theta_{\text{final}} = \Theta' + \Delta\Theta'
\end{equation}

\textbf{Generation Stage}: After completing both offline and online parameter injections, the final model parameters $\Theta_{\text{final}}$ integrate legal knowledge from the entire corpus. The model is now ready to be directly applied to downstream tasks for experimental evaluation.

\section{Experiment Setting}

\subsection{Collection and Annotation}

We construct \textbf{Legal-CA}, a multi-task benchmark comprising 1,990 training instances and 590 test instances. Detailed statistics are presented in Table~\ref{Legal-CA_info}.  
To support legal knowledge retrieval, we also introduce a corpus named \textbf{Legal-KD}, which contains all criminal, administrative, and civil cases published between 2018 and 2021. All documents have been rigorously anonymized to ensure compliance with ethical and privacy standards.
Additionally, three Master of Laws students, each having passed the National Legal Professional Qualification Examination, are invited to assist in the quality evaluation of the collected cases. 

\subsubsection{Parametric Set}

To ensure the structure, authority, and professional quality of our dataset, we collected 4,000 legal documents from PKULaw, each annotated by certified legal professionals. These documents include structured elements such as case descriptions, key issues, legal reasoning, judgment outcomes, and case analyses, as illustrated in Figure~\ref{case_structure}. 
After rigorous filtering based on quality, representativeness, and legal complexity, a total of 1,990 high-quality cases are retained for parametric injection.

\begin{figure}[t]  
	\centering  
	\includegraphics[width=7cm, height=3cm]{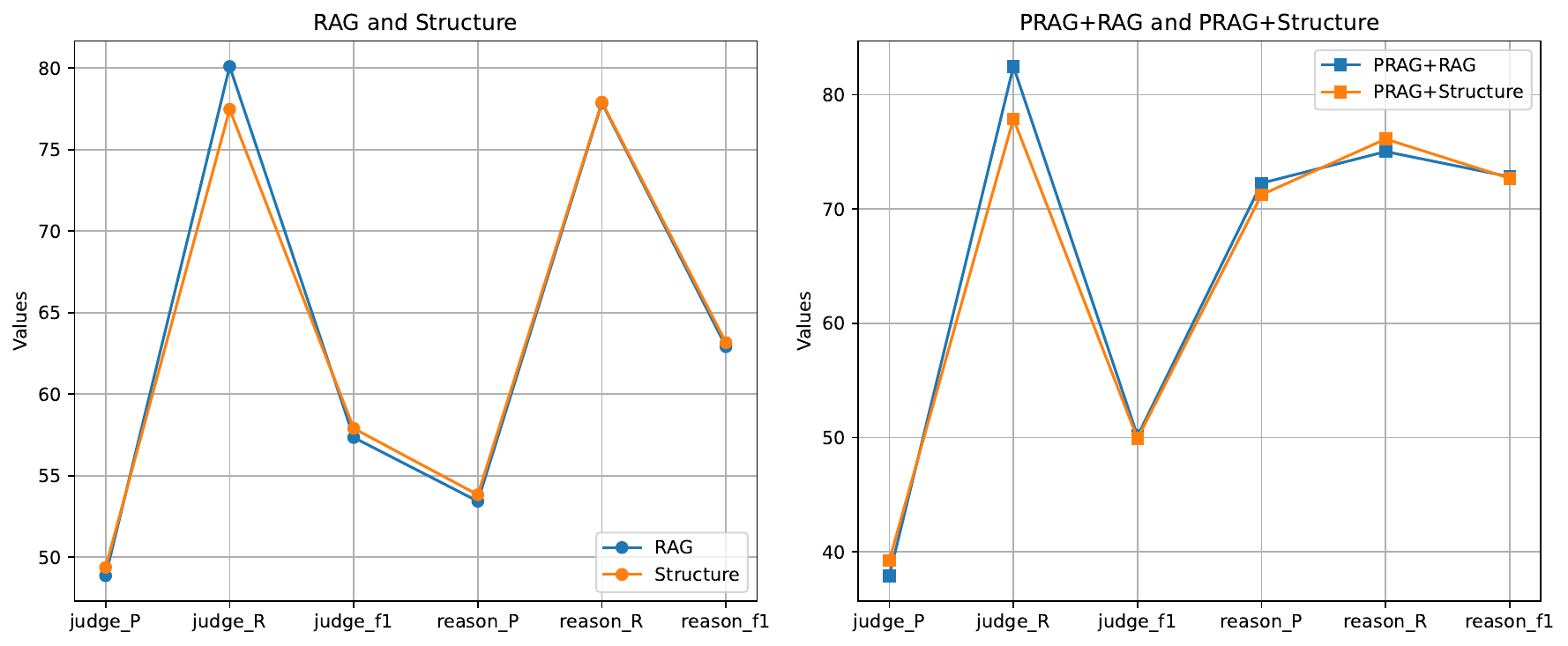}  
	\caption{The performance of plain RAG and structure RAG.}
	\label{structure}  
\end{figure}

\begin{table*}[t]
\centering
\normalsize
\begin{tabular}{lcccccccccc}
\toprule
\textbf{Model} & LA-P & LA-R & LA-F1 & Charge & Imprison & Probation & Fine & CA-P & CA-R & CA-F1  \\
\midrule
Qwen1.8B & 6.82 & 7.13 & 6.97 &  66.67 & 38.80 & 35.36 & 36.39 & 3.26 & 7.85 & 4.87 \\
Qwen1.8B + RAG & \underline{10.58} & \textbf{10.49} & \underline{10.54} &  \underline{81.82} & \underline{40.32} & 38.07 & \textbf{39.66} & 14.87 & 17.80 & 16.21 \\
Qwen1.8B + P-RAG & 10.14 & 6.12 & 7.63 &  73.84 & 40.12 & \underline{38.25} & 39.04 & \textbf{22.42} & \underline{22.68} & \textbf{22.55} \\
Combine Both & \textbf{13.92} & 9.01 & \textbf{10.94} & \textbf{86.91} & \textbf{40.53} & \textbf{39.77} & \underline{39.09} & \underline{21.33} & \textbf{23.56} & \underline{22.39} \\
\midrule
\end{tabular}
\caption{Fine-grained performance of Qwen1.5-1.8B-Chat in Legal-CA framework.}
\label{1.8B}
\end{table*}

\subsubsection{Test Set}

We collected 800 legal cases from the official \textit{Wenshu Court}, covering criminal, administrative, and civil domains. To prevent data leakage, only documents published after January 1, 2025, were selected. Furthermore, to ensure sufficient case complexity, we required the factual descriptions to exceed 200 Chinese characters. After filtering, 590 cases were retained.  
As shown in Table~\ref{datasets_difference}, compared with existing datasets, Legal-CA ranks among the top in terms of average token count and number of relevant articles. Each case is manually annotated by the three law students, who segment the documents into the following components: basic case facts, legal reasoning, judgment outcomes, and relevant articles, consistent with the structure shown in Figure~\ref{case_structure}.

\begin{table*}[t]    
    \centering
    \normalsize
    \begin{tabular}{lcccccccc}
        \toprule
        Model(\%) & LA-F1 & Charge & Imprison & Probation & Fine & CA-F1 & J-F1 & R-F1 \\
        \midrule
        ONLINE & 29.49 & 88.19 & 43.67 & 39.45 & 41.81 & 31.23 & 76.47 & 74.7 \\
        OFFLINE & 16.47 & 84.32 & 42.05 & 40.47 & 40.43 & 19.38 & 69.96 & 72.5 \\
        \bottomrule
    \end{tabular}
    \caption{Performance of online and offline. "LA" means legal article, "CA" means Civil \& Admin, and "R" means Reason.}
    \label{online_offline}
 \end{table*}

\subsubsection{Legal Knowledge Base}

We propose Legal-KD, a legal case knowledge database constructed from the corpus of judicial documents published by the Wenshu Court, encompassing all Chinese legal cases between 2018 and 2021. This corpus serves as the source for online retrieval in our legal case retrieval framework. All documents in the dataset are unstructured plain texts. Given the large volume of judgments, we limit the number of documents per cause of action to a maximum of 10, and exclude any cases containing fewer than 150 Chinese characters to ensure data quality. As shown in Table~\ref{article_recall}, the BM25 retrieval method consistently achieves the best performance; therefore, we adopt BM25 to retrieve relevant cases. Implementation details and further processing procedures are available in our code.

\subsection{Metric}

\subsubsection{LJP}

For criminal law cases, we evaluate the model’s predictions based on charges, prison terms (including imprisonment and probation), and fines. For administrative and civil cases, we focus on key outcomes such as the plaintiff's litigation result, monetary compensation, and ownership disputes.
However, accurately predicting prison terms or fines remains challenging for LLMs. As noted in prior studies~\cite{he2024agentscourtbuildingjudicialdecisionmaking,chang2025asp2ljadversarialselfplay}, criminal case descriptions often include the prosecution's sentencing recommendations, allowing models to simply copy this information to achieve high accuracy. This shortcut fails to reflect the model’s true reasoning ability.
To ensure a fair evaluation of generated content, we remove the prosecution's claims from the case fact descriptions and introduce a new evaluation metric, defined as follows (where $d$ denotes the difference, $r$ the reference answer, and $h$ the hypothesis):

\begin{equation}
d(r, h) = 1 - \frac{1}{1 + \exp\left(-\frac{|r - h|}{|r| + |h| + \epsilon}\right)}
\label{eq:custom_loss}
\end{equation}

\subsubsection{SAG}

Similar to SAR, SAG focus on automatically generating the most relevant legal articles based on case descriptions and retrieved documents. Given an input query, the LLM is required to generate relevant legal articles and determine whether the generated provisions are encompassed within the correct set of legal articles.

\subsubsection{LDG}

This task not only requires correct judgment outcomes but also emphasizes the interpretability and legal validity of the reasoning process. Models need to produce human-like documents based on basic case facts.
To assess the model’s legal language generation capabilities, we focus on the semantic alignment between generated and ground-truth legal documents. Specifically, we adopt the \texttt{chinese-roberta-wwm-ext} model~\cite{chinese-bert-wwm} to evaluate semantic relevance, thereby reflecting the model’s effectiveness in legal expression.

\begin{table}[t]
\centering
\scalebox{0.65}{
    \begin{tabular}{lcccccc} 
    \toprule
    \textbf{Features} & \textbf{Crime-1} & \textbf{Admin-1} & \textbf{Civil-1} & \textbf{Crime-2} & \textbf{Admin-2} & \textbf{Civil-2} \\
    \midrule
    cases & 192 & 203 & 196 & 697 & 266 & 1027 \\ 
    average articles & 6.7 & 7.0 & 6.5 & 2.6 & 2.8 & 3.4 \\
    max articles & 15 & 23 & 21 & 18 & 12 & 20 \\
    total articles & 1280 & 1415 & 1271 & 1821 & 745 & 3457\\
    avg len per case fact & 440.9 & 384.7 & 482.1 & 430.3 & 560.7 & 700.0 \\
    \bottomrule
    \end{tabular}
}
\caption{Comparison with other datasets. The suffix "-1" means test set, and "-2" means train set.}
\label{Legal-CA_info}
\end{table}

\subsection{Baseline}
We choose several general or legal models as our baselines:
\begin{itemize}
    \item \textbf{Vanilla}: We adopt Qwen1.5-7B-Chat as the base model. For comparison, we also include GPT-3.5-turbo-0125, GPT-4o-mini, and GPT-4o-2024-11-20.
    \item \textbf{LexiLaw}~\cite{LexiLaw}: A model fine-tuned on large-scale Chinese legal corpora, demonstrating strong legal knowledge and reasoning capabilities. It is based on ChatGLM-6B.
    \item \textbf{ChatLaw}~\cite{cui2024chatlaw}: We adopt ChatLaw-13B, built upon Ziya-LLaMA-13B-v1, which performs well across various Chinese legal tasks.
\end{itemize}

\subsection{Hyperparameters}
For all models, the temperature is set to 0.7.  
For LoRA fine-tuning, we set the \texttt{learning\_rate} to $1 \times 10^{-5}$, \texttt{lora\_rank} to 2, \texttt{lora\_alpha} to 32, and the number of training epochs to 1. Further implementation details are provided in our code.

\section{Results and Analysis}

\subsection{Main Results}

In this section, we conduct a comprehensive experiment and present its results in Table~\ref{fine-grained} and~\ref{coarse-grained}. We categorize the models into four groups: base models, vanilla RAG, P-RAG, and the paradigm combining vanilla with P-RAG. The evaluation is conducted across both coarse-grained and fine-grained tasks. Our key findings are as follows:

(1) P-RAG exhibits a positive effect in enhancing model capabilities. Compared with GPT-4o, which is currently the most powerful model, our PL-CA, built upon Qwen1.5-7B-Chat, achieves overall superior performance. Notably, in the SAG task, the F1 score of PL-CA exceeds that of GPT-4o by approximately 8.5 points. On more challenging tasks, such as predicting prison terms and amounts, PL-CA also outperforms all baselines except GPT-4o. These results demonstrate that parameter-level knowledge injection can significantly improve model capabilities, offering valuable insights for the design of future knowledge integration methods.

(2) To better compare vanilla RAG and P-RAG, we present the results of the base model equipped with RAG in the Table~\ref{fine-grained}. P-RAG consistently outperforms vanilla RAG, suggesting that LLMs are more effective at utilizing parametric knowledge than context-based information. This observation aligns with prior findings in~\cite{yu2024neuronlevelknowledgeattributionlarge}.

\subsection{Ablation}
\subsubsection{Structure}
As shown in Table~\ref{coarse-grained}, we observe that Qwen's performance on coarse-grained metrics declined after incorporating retrieved contextual information, regardless of whether P-RAG is applied. We hypothesize that this performance degradation may stem from the unstructured nature of the retrieved legal documents, which are stored as free-text strings without explicit annotation or segmentation.

To investigate this hypothesis, we reprocessed the retrieved corpus into a structured format, explicitly annotating key legal elements such as case facts, legal reasoning, judgments, and relevant legal articles. These structured components were then incorporated into the model's input context. The corresponding experimental results are presented in Figure~\ref{structure}.

Interestingly, we found that the performance improvements achieved through structured retrieval documents were comparable to those obtained using unstructured ones. This finding suggests that simply injecting structured content into the model's context does not substantially enhance its semantic understanding or representation capability. To more effectively align the model's output with human-like legal reasoning, alternative strategies such as supervised fine-tuning or reinforcement learning may be necessary. We leave the exploration of these directions to future work.

\subsection{Offline and Online}
To better measure the effects of online and offline learning, we trained Qwen1.5-7B separately using only online and offline. To obtain more objective experimental results, we reduced the amount of offline learning data to 600 samples, making it approximately equal to the amount of online learning data. The results are shown in the table. It is indicated that both online learning and offline learning have positive effects on the model. Due to the higher semantic relevance between the query and the retrieved online cases, online learning proves more effective than offline learning under equal data scale conditions. Therefore, online learning results in greater performance improvements.

\subsection{Parameter Scale}
Furthermore, we selected the smaller model Qwen1.5-1.8B-Chat to investigate whether P-RAG could enhance the capabilities of small models. As shown in Table~\ref{1.8B}, although the 1.8B model has limited capabilities, P-RAG still leads to improvements.
Interestingly, we observe that while P-RAG brings improvements to the 1.8B model, its performance gains are less obvious compared to traditional RAG and the combined approach, which contrasts with the findings for the 7B model. This suggests that P-RAG is more effective for larger-scale models, whereas smaller models benefit more from direct context injection. This observation calls for further qualitative and quantitative analysis.

\section{Conclusion}
In this work, we propose \textbf{PL-CA}, a novel parametric legal case augmentation framework that integrates LLM with parametric knowledge injection to mitigate the limitations of conventional context-based retrieval in legal AI. 
To systematically evaluate the performance of LLMs across a broad spectrum of legal tasks, we also construct \textbf{Legal-CA}, a comprehensive benchmark comprising both coarse-grained and fine-grained subtasks. Legal-CA reflects real-world legal challenges such as sentencing prediction, dispute focus identification, and structured judgment generation.

Extensive experiments demonstrate that our method, based on Qwen1.5-7B-Chat, achieves superior overall performance compared to traditional RAG methods, as well as powerful closed-source models like GPT-4o. Our findings highlight the effectiveness of parametric injection in enhancing the legal capabilities of LLMs, especially on tasks involving complex legal semantics. This work offers valuable insights for future research in developing scalable, high-performance legal AI systems.

\bibliography{main}


\end{document}